\pdfoutput=1
\documentclass[10pt,twocolumn,letterpaper]{article}

\usepackage[pagenumbers]{wacv}

\usepackage{tabularx}
\usepackage{booktabs}
\usepackage[table]{xcolor}
\usepackage{cuted}

\definecolor{rankFirst} {rgb}{1.00, 0.70, 0.70}
\definecolor{rankSecond}{rgb}{1.00, 0.85, 0.70}
\definecolor{rankThird} {rgb}{1.00, 1.00, 0.70}

\newcommand{\swatch}[1]{%
  {\setlength{\fboxsep}{0pt}%
   \colorbox{#1}{\rule{0.7em}{0pt}\rule{0pt}{0.7em}}}%
}

\newcommand{\first}[1]{\cellcolor{rankFirst}#1}
\newcommand{\second}[1]{\cellcolor{rankSecond}#1}
\newcommand{\third}[1]{\cellcolor{rankThird}#1}

\definecolor{wacvblue}{rgb}{0.21,0.49,0.74}
\usepackage[pagebackref,breaklinks,colorlinks,allcolors=wacvblue]{hyperref}

\def\wacvPaperID{3254}
\def\confName{WACV}
\def\confYear{2027}

\title{Field2Sim: Turning In-Field Images into Simulation-Ready Plant Assets}

\author{Diram Tabaa \ \ \ \
Gianni Di Caro\\
Carnegie Mellon University\\
Doha, Qatar\\
{\tt\small \{dtabaa, gdicaro\}@andrew.cmu.edu}
}

\begin{document}
\maketitle

\begin{strip}
    \centering
    \vspace{-3.5em}

    \includegraphics[
        width=\textwidth,
    ]{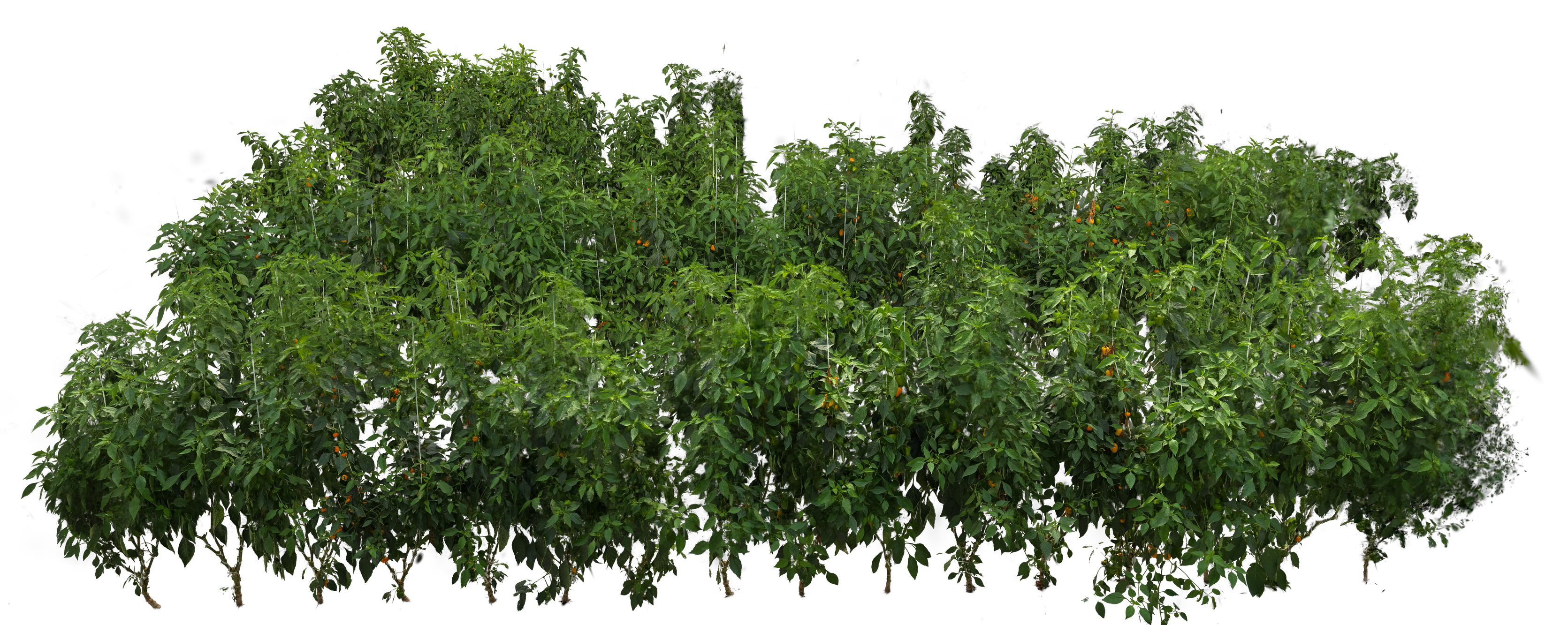}

    \refstepcounter{figure}
    \label{fig:teaser}
    \vspace{0.3em}

    \begin{minipage}{0.96\textwidth}
        \small
\textbf{A row-scale plant asset reconstructed by Field2Sim.}
From handheld smartphone images, Field2Sim recovers a photorealistic
Gaussian representation of a fruit-bearing capsicum row and separates
it from the surrounding environment. The resulting plant-only asset
preserves the appearance and structural variation of the observed
crop and is further decomposed into individual fruit instances for
visual simulation and perception.
    \end{minipage}

    \vspace{-0.6em}
\end{strip}
\begin{abstract}
 Realistic, semantically annotated plant models are essential for evaluating agricultural robot perception and view planning under representative visual conditions in simulation, but creating such assets from real crops remains difficult. Existing approaches often trade annotation quality for visual realism or require specialized capture equipment. We introduce Field2Sim, a modular pipeline that converts handheld smartphone images of crop rows into photorealistic Gaussian Splatting reconstructions with separated plants and individually identified fruits. Field2Sim selects its reconstruction backbone based on downstream fruit-segmentation performance, transfers 2D semantic masks into 3D without additional segmentation training, and introduces a novel two-pass fruit-instance identification method that refines geometric clusters using multi-view 2D instance evidence. We evaluate Field2Sim on five capsicum rows in a commercial greenhouse containing 1,722 annotated fruits, where the two-pass method achieves a 3D panoptic quality of 0.954, compared with 0.808 for geometric clustering alone, and reduces fruit-count error from 20.0\% to 1.5\%. We publicly release a dataset of the five reconstructions with their plant and fruit annotations, together with our code, to support simulation-based research on agricultural robot perception and view planning.
\end{abstract}\vspace{-0.5em}

\section{Introduction}
\label{sec:introduction}

Agricultural automation, from field robots to networked camera and sensor systems, increasingly relies on simulated environments to train and validate its perception and decision-making components~\cite{Shamshiri_2018}.
These environments can provide perception models with labeled training images, such as segmentation masks and depth maps, without requiring manual annotation \cite{martini_enhancing_2024}.
They also allow perception-driven robot policies, such as view planning or navigation, and different sensor configurations to be evaluated in closed loop prior to deployment in real fields, including from viewpoints that need not have been physically captured \cite{fujinaga_autonomous_2025}.
Since plants make up most of what these systems observe, the utility of such environments depends on the visual realism of their 3D plant models and on the semantic information those models provide.
Existing approaches to generating crop assets typically achieve one of these properties at the expense of the other.
Procedural systems and artist-designed meshes provide complete semantic labels by construction, along with controllable geometry and straightforward scene randomization~\cite{10.3389/fpls.2019.01185, cropcraft}.
However, their realism is limited by the complexity of the underlying models or the manual effort invested in their design, and the resulting assets can remain visibly synthetic.
Reconstruction from real plants addresses this limitation.
Terrestrial laser scanners, structured-light sensors, and X-ray tomography can recover plant geometry with high accuracy~\cite{10.1371/journal.pone.0256340, Dutagaci2020ROSEXAA}, but require specialized acquisition equipment and do not inherently provide photorealistic appearance or semantic labels.
Image-based reconstruction using radiance fields or Gaussian splatting lowers the barrier to acquisition by relying on ordinary cameras rather than specialized sensing equipment while retaining detailed visual appearance.
In plant applications, however, these representations have largely been developed for phenotyping, with semantic or instance information used primarily to support measurements and analysis rather than to produce reusable, semantically annotated plant assets~\cite{Zhang_Gajardo_2025_CVPR, fruitnerf2024, fruitnerfpp2025}.
As such, obtaining plant assets that are both photorealistic and semantically annotated from real crops, using only commodity cameras, remains difficult.

To fill this gap, we propose \textbf{Field2Sim}, a modular pipeline built around Gaussian Splatting \cite{kerbl3Dgaussians} that generates simulation-ready plant assets from images captured in fields and greenhouses.
We define a plant asset as \emph{simulation-ready} if it (i) preserves a photorealistic appearance, (ii) encodes semantic information, and (iii) can be readily imported into existing simulation frameworks.
Field2Sim leverages recent advances in vision foundation models, such as SAM~3~\cite{carion2025sam3segmentconcepts}, to generate 2D semantic masks for the captured images; these masks are then lifted onto the 3D Gaussian primitives using training-free techniques.
Importantly, we validate our choice of Gaussian Splatting training method through its downstream semantic segmentation performance, since standard photometric metrics (e.g., PSNR, SSIM, and LPIPS) do not capture the particular challenges posed by agricultural scenes.
Finally, we introduce a novel two-pass instance segmentation method that addresses the difficulty of assigning individual instance labels to densely clustered and partially occluded fruits, and we compare it against geometry-based approaches.

In summary, our main contributions are:
\begin{itemize}
    \item \textbf{Field2Sim}, a modular pipeline that transforms casually captured greenhouse images into photorealistic, semantically decomposed plant and fruit assets, without specialized sensing equipment or scene-specific fine-tuning.
    \item A \textbf{controlled evaluation} of the pipeline's main design choices, comparing five Gaussian Splatting reconstruction strategies by downstream semantic segmentation, and a novel \textbf{two-pass fruit instance segmentation} method evaluated against geometric clustering baselines.
    \item A \textbf{curated dataset} of five capsicum greenhouse-row reconstructions with semantic plant representations and individual fruit assets, publicly released\footnote{\url{https://dtabaa.github.io/field2sim}} to support perception and planning research in agricultural vision and robotics.
\end{itemize}

\section{Related Work}
\label{sec:related_work}

\par\noindent
\textbf{Radiance Field Reconstruction.}\enspace Neural Radiance Fields (NeRFs)~\cite{mildenhall2020nerf}
demonstrated photorealistic novel-view synthesis by optimizing
a continuous scene representation through differentiable volume
rendering. 3D Gaussian Splatting (3DGS)~\cite{kerbl3Dgaussians}
subsequently introduced an explicit representation based on
anisotropic Gaussian primitives, enabling faster optimization,
real-time rendering, and direct manipulation of the reconstructed
scene. Several extensions address limitations of its original
optimization strategy: \mbox{AbsGS}~\cite{ye2024absgs} mitigates gradient
cancellation during densification, Taming 3DGS~\cite{10.1145/3680528.3687694}
controls primitive allocation under a fixed budget, and
3DGS-MCMC~\cite{kheradmand20243d} replaces heuristic densification
with a sampling-based formulation. Complementary approaches improve
scale-aware rendering~\cite{Yu2024MipSplatting} or support large-scale scenes
through hierarchical optimization~\cite{hierarchicalgaussians24}.
While these methods perform strongly on standard benchmarks, their
relative behavior remains underexplored for row-grown crops, which
combine repeated appearance, thin structures, severe occlusion, and
non-uniform view coverage.

\par\noindent
\textbf{3D Scene Understanding.}\enspace While 3D scene understanding has traditionally operated on point
clouds and meshes, differentiable rendering enables semantic
information from pretrained 2D models to be transferred directly
into radiance-field representations. LERF~\cite{lerf2023}
distills CLIP features into a NeRF, while Feature
3DGS~\cite{zhou2024feature} and
LangSplat~\cite{qin2024langsplat} attach distilled features to
Gaussian primitives for efficient open-vocabulary querying.
Rather than learning an additional feature field,
FlashSplat~\cite{flashsplat} directly solves the assignment
of 2D masks to Gaussians, whereas LUDVIG~\cite{marrie2025ludvig}
uses learning-free feature aggregation and graph diffusion.
Instance-level understanding remains more involved because 2D
instance identities are generally inconsistent across views.
Methods such as \mbox{Panoptic Lifting}~\cite{Siddiqui_2023_CVPR},
GARField~\cite{garfield2024}, and
\mbox{PanopticSplatting}~\cite{11245846} address this
through learned instance labels, affinity fields, or object
queries. For our application, we instead employ a post-hoc
procedure that combines geometric grouping with 2D instance
evidence to recover individual fruits.

\par\noindent
\textbf{Plant Reconstruction and Simulation.}\enspace
Agricultural simulation has traditionally relied on procedural plant models. Helios~\cite{10.3389/fpls.2019.01185} combines organ-level geometry with biophysical simulation, while CropCraft~\cite{cropcraft} generates configurable crop fields and annotations for agricultural perception. These systems provide controllable geometry and labels, but their structural diversity and appearance remain governed by authored models and procedural rules.

Image-based methods preserve observed crop morphology, although most target specific phenotyping outputs. Choi \etal~\cite{10.3389/fpls.2024.1439086} extract NeRF geometry and apply skeletonization, surface reconstruction, and geometric fitting to estimate plant traits. FruitNeRF~\cite{fruitnerf2024} samples a semantic NeRF into a fruit point cloud for clustering, while FruitNeRF++~\cite{fruitnerfpp2025} introduces contrastive instance embeddings for fruit-agnostic counting. CountingFruit~\cite{li2025countingfruit} instead embeds language features into 3D Gaussians and clusters prompt-selected samples, and Wheat3DGS~\cite{Zhang_Gajardo_2025_CVPR} associates detector- and SAM-derived masks with Gaussians to isolate and measure wheat heads. These methods primarily produce task-specific counts or measurements rather than reusable plant and fruit components. GrowSplat~\cite{adebola2025growsplat} and Botany-Bot~\cite{adebola2025botanybot} reconstruct richer plant digital twins, but rely on fixed multi-camera or controlled robotic acquisition. In contrast, Field2Sim targets simulation-ready reconstruction from more accessible data: it reconstructs greenhouse rows from handheld images, then applies post-hoc semantic lifting and instance separation to obtain plant- and fruit-level assets without dedicated scanning infrastructure.

\section{Field2Sim}
\label{sec:field2sim}

We present Field2Sim, a framework for generating
simulation-ready agricultural assets from images captured with
commodity hardware. This section outlines the pipeline (\cref{sec:system_overview}) and the principles behind its design (\cref{sec:design_principles}).

\begin{figure*}[t]
\centering
\includegraphics[width=\textwidth]{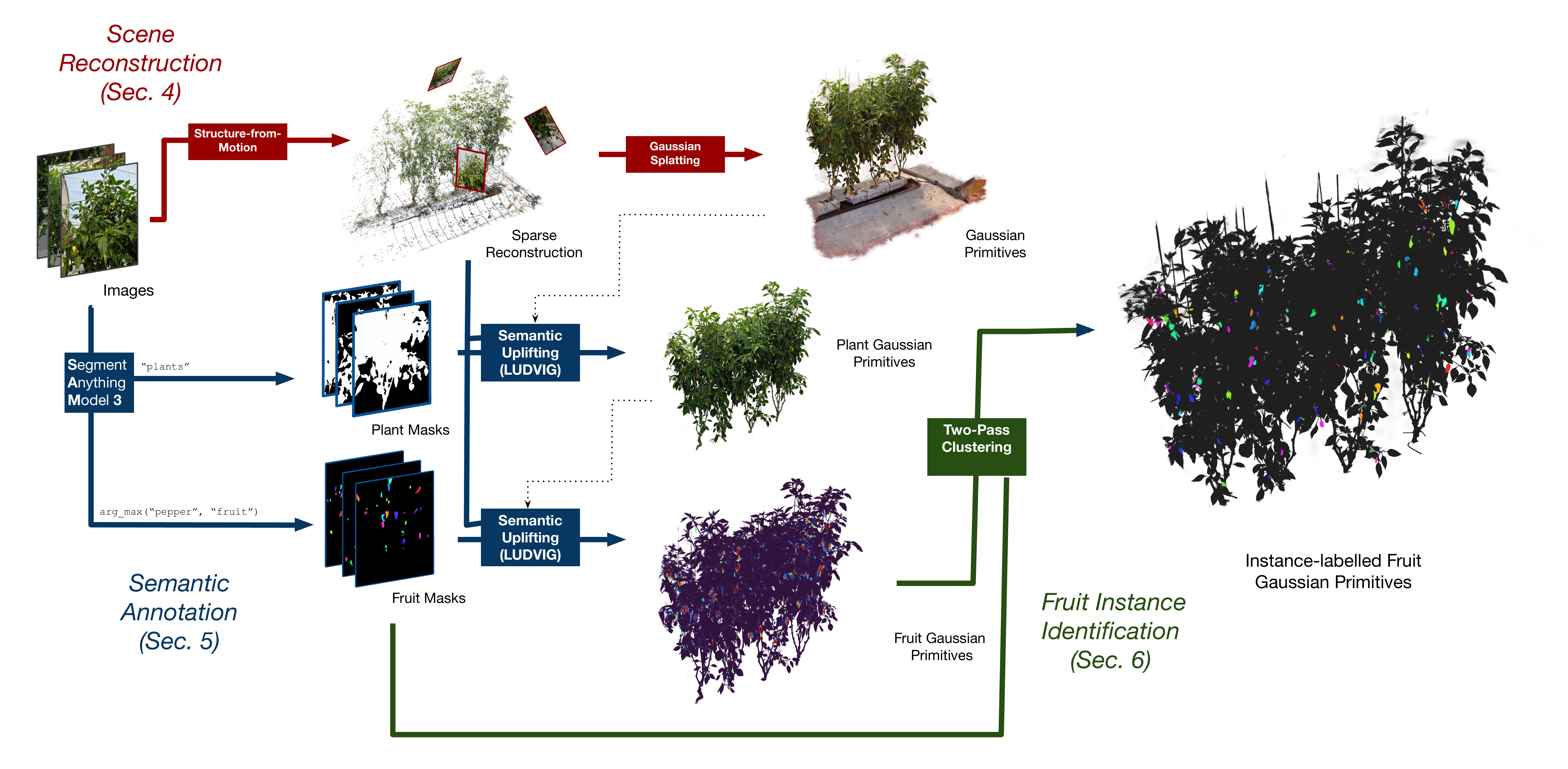}
\caption{\textbf{Field2Sim pipeline.}
Handheld RGB images are reconstructed into 3D Gaussian
primitives, semantically decomposed into plant and fruit
representations, and processed to identify individual fruit
instances. Dotted arrows indicate reconstruction information
reused during semantic lifting.}
\label{fig:pipeline}
\end{figure*}

\subsection{System Overview}
\label{sec:system_overview}

Given a set of handheld RGB images of a crop row, Field2Sim
constructs a photorealistic 3D Gaussian representation with
plant-level semantic labels and individual fruit instances.
As illustrated in Fig.~\ref{fig:pipeline}, we first estimate
camera poses and sparse scene geometry using \mbox{Structure-from-Motion}
(SfM), then use these estimates to optimize a 3D Gaussian
Splatting (3DGS) representation. The reconstructed scene is
subsequently aligned and spatially cropped to remove distant
artifacts. We apply SAM~3 to obtain semantic plant masks and
fruit instance masks, which \mbox{LUDVIG} lifts into 3D in two
successive passes to isolate plant and fruit Gaussians.
Individual fruits are then identified through a two-stage
procedure: an initial decomposition into 3D connected
components, followed by selective refinement of components
containing evidence of multiple fruit instances in the 2D
masks. The resulting representation preserves the appearance
of the captured crop row while exposing individual plants
and fruits for downstream simulation tasks.

\subsection{Design Principles}
\label{sec:design_principles}
We follow three design principles that guide both the selection of
existing components and the design of our own methods within the
framework:

\begin{itemize}
    \item \textbf{Modularity.}
    Each stage makes minimal assumptions about adjacent components,
    allowing individual methods to be replaced as stronger alternatives
    become available.

    \item \textbf{No fine-tuning.}
    We favor methods that operate out of the box, avoiding
    dataset- or scene-specific training and additional annotations.

    \item \textbf{Scalability.}
    We prioritize methods that require limited manual intervention and
    remain practical as the number of images, plants, and scenes grows.
\end{itemize}

The following sections detail the pipeline in order.
We evaluate the component choices and
resulting assets in Sec.~\ref{sec:experiments}.

\section{Scene Reconstruction}
\label{sec:reconstruction}

\subsection{Sparse Geometry Estimation}

Our framework uses Gaussian Splatting to obtain photorealistic
representations of agricultural plants, requiring camera poses
and a sparse 3D point cloud to initialize the Gaussian primitives.
We recover both using an incremental Structure-from-Motion
(SfM) pipeline~\cite{schoenberger2016sfm},
which establishes feature correspondences across images and
jointly estimates camera poses and sparse scene geometry.
Since reconstruction is performed offline, we prioritize
reconstruction robustness over computational efficiency.

The strong visual self-similarity between plants and repeated
structures can produce incorrect correspondences between
spatially distinct regions, degrading image registration and
pose estimation. We therefore restrict feature matching to a
sliding window of 50 images around each image in the acquisition
sequence. Unlike exhaustive or vocabulary-tree-based matching,
this constraint uses the acquisition order to reduce
geometrically implausible correspondences between visually
similar plant regions.

\subsection{Radiance Field Optimization}
\label{sec:radiance_field_optimization}

Given the recovered camera poses and sparse geometry, we optimize the scene as a set of 3D Gaussian primitives. We use 3DGS because its representation is explicit, portable, and can be directly rendered and edited by downstream stages. We therefore avoid variants that introduce additional neural parameters, such as Scaffold-GS~\cite{scaffoldgs}, which would make the exported representation less self-contained.

Based on the empirical comparison presented in Sec.~\ref{sec:reconstruction_comparison}, we use the chunk-level optimizer of Hierarchical 3DGS~\cite{hierarchicalgaussians24}. Its densification strategy replaces the average screen-space positional gradient used by the original 3DGS with the maximum gradient observed over the densification window. Let \(\tilde{\boldsymbol{\mu}}_i^{(t)}\) denote the image-space mean of Gaussian \(i\) at optimization step \(t\), and let \(\mathcal{T}_i\) contain the steps during which it is observed. The corresponding densification statistic is

\begin{equation}
g_i =
\max_{t \in \mathcal{T}_i}
\left|
\frac{\partial \mathcal{L}_t}
{\partial \tilde{\boldsymbol{\mu}}_i^{(t)}}
\right|_2 ,
\end{equation}

with Gaussian \(i\) densified when \(g_i > \tau_{\mathrm{dens}}\). This prevents strong refinement signals from a small number of informative views from being diluted by weaker observations, which is useful under the uneven view coverage and heavy occlusion present in our captures. We use only this chunk-level optimization and do not construct the global hierarchy, leaving a standard set of Gaussian primitives for subsequent processing.

\section{Semantic Annotation}
\label{sec:semantic}

Several methods augment Gaussian splatting with semantic
information~\cite{zhou2024feature,qin2024langsplat,flashsplat}.
However, many require optimizing additional scene-specific features
or coupling semantic lifting with reconstruction. To preserve the
modularity of Field2Sim, we instead use
LUDVIG~\cite{marrie2025ludvig}, which transfers 2D masks to an
already optimized Gaussian representation without further training
or modifying its parameters.

We apply semantic lifting sequentially. Plant masks first separate
the crop from the surrounding greenhouse structures. Fruit masks
are then lifted onto the resulting plant representation to isolate
the fruit Gaussians used for instance identification. Although we
focus on peppers, the same procedure can be applied to other visible
plant organs, such as flowers or leaves, by changing the 2D prompt.

\subsection{2D Semantic Mask Generation}

We use SAM~3~\cite{carion2025sam3segmentconcepts} because it performs
well in our setting without fine-tuning. For every input image, we
generate a semantic plant mask and both semantic and instance-level
fruit masks. The semantic masks provide supervision for 3D lifting,
whereas the instance masks are retained as multi-view evidence for
the fruit-separation stage described in
Sec.~\ref{sec:instances}.

Since SAM~3 is text-prompted, its predictions depend on the noun
used to describe the target. The prompt \texttt{plants} is used for
plant isolation across all scenes. For fruit segmentation, we
compare a small set of candidate nouns on sampled frames from each
scene and select the prompt producing the strongest detection
coverage.

\subsection{3D Semantic Uplifting}
\label{subsec:uplifting}

LUDVIG interprets semantic lifting as an inverse rendering
operation. Let \(q=(d,p)\) denote pixel \(p\) in view \(d\), and let
\(\mathcal{S}_i\) contain all such observations influenced by
Gaussian \(i\). Its uplifted semantic score is

\begin{equation}
s_i =
\frac{\sum_{q\in\mathcal{S}_i} w_{iq} M_q}
     {\sum_{q\in\mathcal{S}_i} w_{iq}},
\end{equation}

where \(M_q\) is the corresponding binary mask value and \(w_{iq}\)
is the alpha-blending contribution of Gaussian \(i\) to that pixel.
Thus, \(s_i\) measures how consistently the Gaussian contributes to
foreground pixels across the observed views.

We first uplift the plant masks onto the complete scene and retain
Gaussians whose scores exceed
\(\tau_{\mathrm{plant}}=0.3\), producing a plant-only
representation. We then repeat the operation on this filtered
representation using the semantic pepper masks and
\(\tau_{\mathrm{fruit}}=0.5\). The result is a fruit-only set of
Gaussians, while their geometry, appearance, opacity, and other
reconstruction parameters remain unchanged. These Gaussians form
the input to the instance-identification procedure described in
Sec.~\ref{sec:instances}.

\section{Fruit Instance Identification}
\label{sec:instances}

Agricultural simulation tasks such as fruit counting and the evaluation of view-planning coverage require instance masks that distinguish individual fruits. Separating fruit instances also allows plant configurations to be randomized by removing, duplicating, or repositioning fruits (see \cref{sec:practical_utility}). However, tightly clustered or touching fruits can form a single connected structure in 3D, making it difficult to recover individual instances from the fruit-level semantic representation. Occlusion and incomplete observations further limit the reliability of shape- and size-based separation. We therefore propose a two-pass procedure that first groups fruit Gaussians by geometric proximity, then uses multi-view instance masks to identify and refine components containing multiple fruits.

\par\noindent
\textbf{Initial Pass.}\enspace
We first construct a proximity graph over the segmented fruit Gaussians and
extract its connected components. Since Gaussian primitives encode both position
and spatial extent, we use a symmetric Mahalanobis distance rather than Euclidean
distance. For a pair of Gaussians $i$ and $j$, with means $\boldsymbol{\mu}_i$
and $\boldsymbol{\mu}_j$ and covariances $\boldsymbol{\Sigma}_i$ and
$\boldsymbol{\Sigma}_j$, we define
\begin{equation}
d(i,j) =
\sqrt{
(\boldsymbol{\mu}_j-\boldsymbol{\mu}_i)^\top
\left(
\frac{\boldsymbol{\Sigma}_i+\boldsymbol{\Sigma}_j}{2}
+\epsilon\mathbf{I}
\right)^{-1}
(\boldsymbol{\mu}_j-\boldsymbol{\mu}_i)
}.
\end{equation}
To avoid considering all Gaussian pairs, we first generate candidate neighbours
within a Euclidean radius determined by the $99^\mathrm{th}$ percentile of the
maximum Gaussian scales, and retain an edge whenever $d(i,j)<\tau$, where \(\tau\) controls the permissiveness of the initial geometric clustering.

\par\noindent
\textbf{Further Clustering.}\enspace
Because we already obtain per-image fruit instance masks from SAM~3, we use them
to identify components that potentially over-aggregate multiple fruits. For each
view, we render the full scene and recover the standard 3DGS contribution
$w_{i,p}=\alpha_{i,p}T_{i,p}$ of Gaussian $i$ to pixel $p$. This naturally
accounts for occlusion, as Gaussians hidden by other scene elements have
negligible contribution.

Let $\mathcal{G}_k$ denote the
Gaussians belonging to component $k$, and let $\Omega_m$ denote the set of pixels
belonging to instance mask $m$. We measure the association between component
$k$ and mask $m$ by accumulating their contributions,
\begin{equation}
W_{k,m}
=
\sum_{i\in \mathcal{G}_k}
\sum_{p\in \Omega_m}
w_{i,p}.
\end{equation}
We then normalize these associations over all foreground masks,
\begin{equation}
\hat{W}_{k,m}
=
\frac{W_{k,m}}
{\sum_{m'} W_{k,m'}},
\end{equation}
such that $\hat{W}_{k,m}$ represents the fraction of the visible component associated
with instance $m$. A view provides evidence that component $k$ contains multiple
fruits when
\begin{equation}
\left|\{m : \hat{W}_{k,m} \geq \gamma\}\right| \geq 2,
\end{equation}
where $\gamma$ is the minimum fraction of the component's foreground contribution
that a mask must explain to count as a distinct instance.

Finally, let $\mathcal{V}_k$ be the set of views satisfying this criterion for
component $k$. We flag the component for further clustering when
\(
|\mathcal{V}_k| \geq \nu,
\)
where $\nu$ is the minimum number of views that must support the split. This
multi-view requirement reduces sensitivity to isolated 2D segmentation errors.

For each flagged component, we select the views in $\mathcal{V}_k$ and run
SAM~3 in sequence mode, obtaining consistent instance identities across them.
We then uplift these tracked masks onto the component Gaussians using the
procedure described in Sec.~\ref{subsec:uplifting}. If $F_{i,o}$
denotes the resulting association between Gaussian $i$ and tracked instance
$o$, we assign
\begin{equation}
\ell_i = \arg\max_o F_{i,o}.
\end{equation}
Gaussians receiving insufficient evidence inherit the label of their nearest
assigned neighbour, and small resulting fragments are similarly reabsorbed.
The component is split only if at least two valid sub-instances remain after
this assignment; otherwise, it is left unchanged.

We perform this refinement independently for each flagged component rather than
tracking all fruits jointly, substantially reducing the memory requirements of
SAM~3 as the number of simultaneously tracked objects increases.

\section{Experiments}
\label{sec:experiments}

\subsection{Experimental Setup}
\label{sec:experimental_setup}

\begin{table}[t]
    \centering
    \caption{Statistics of the greenhouse scenes used for evaluation.
    All scenes contain sweet pepper (\textit{Capsicum annuum} L.).}
    \label{tab:dataset}
    \small
    \setlength{\tabcolsep}{5pt}
    \renewcommand{\arraystretch}{1.04}

    \begin{tabular}{@{}lcccc@{}}
        \toprule
        & \multicolumn{1}{c}{Capture} &
        \multicolumn{3}{c}{Scene characteristics} \\
        \cmidrule(lr){2-2}
        \cmidrule(l){3-5}
        Scene & Images & Plants & Fruits$^\dagger$ &
        Row length$^\dagger$ (m) \\
        \midrule

        \multicolumn{5}{@{}l}{\textit{Orange-fruited scenes}} \\
        \quad \textsc{Orange-L1} & 1944 & 15 & 438 & 6.4 \\
        \quad \textsc{Orange-L2} & 3681 & 15 & 690 & 6.5 \\
        \addlinespace[2pt]

        \multicolumn{5}{@{}l}{\textit{Red-fruited scenes}} \\
        \quad \textsc{Red-L1} & 4087 & 15 & 250 & 6.4 \\
        \quad \textsc{Red-S1} & 1548 & 6 & 79 & 2.5 \\
        \addlinespace[2pt]

        \multicolumn{5}{@{}l}{\textit{Yellow-fruited scenes}} \\
        \quad \textsc{Yellow-S1} & 875 & 4 & 265 & 1.6 \\
        \midrule
        \textbf{Total}
            & \textbf{12135}
            & \textbf{55}
            & \textbf{1722}
            & \textbf{23.4} \\
        \bottomrule
        \bottomrule
        \multicolumn{5}{@{}l}{
            \footnotesize $^\dagger$Approximate value.
        }

    \end{tabular}
\end{table}

\par
\begin{figure}[t]
    \centering
    \includegraphics[
        width=\columnwidth
    ]{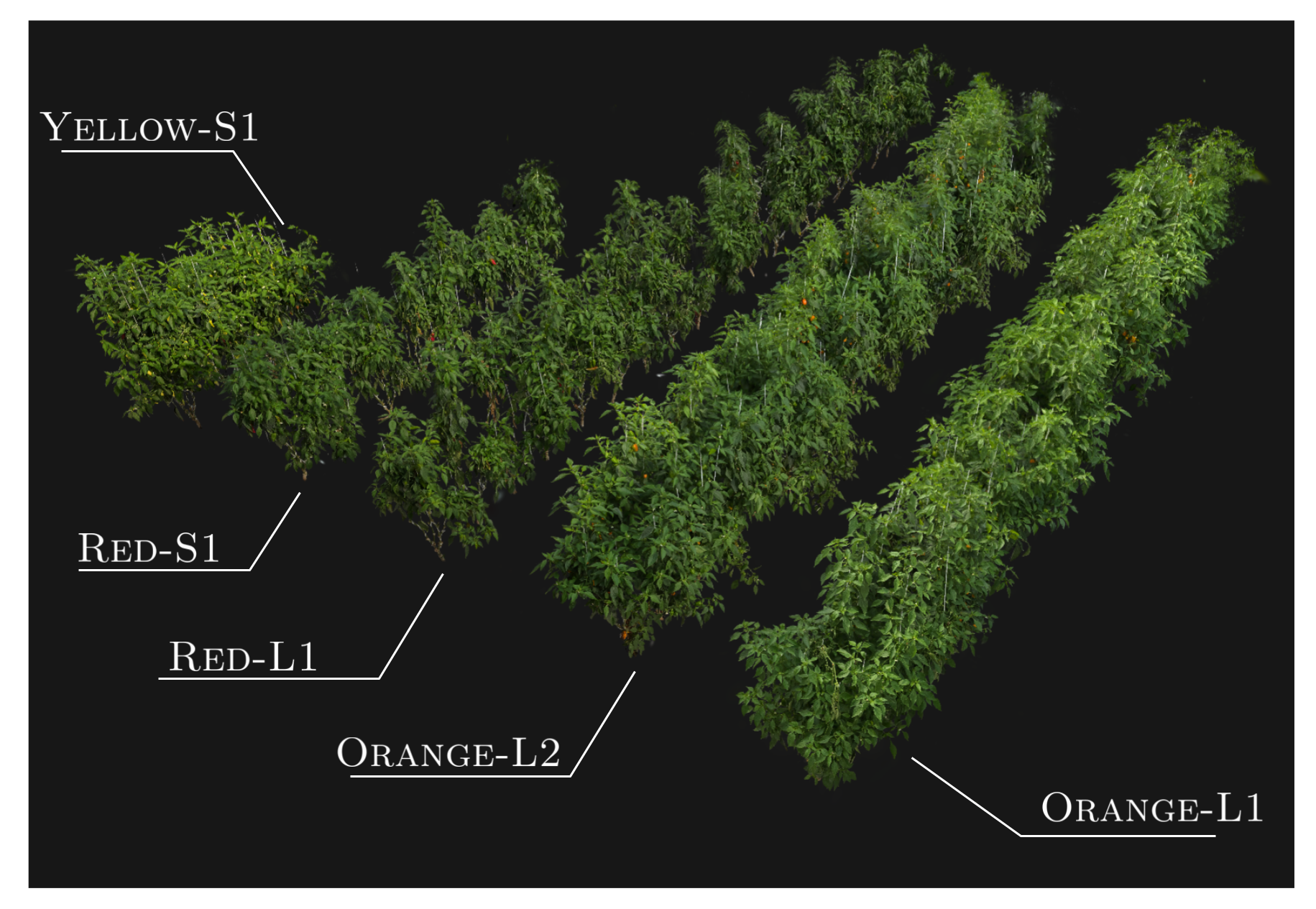}
    \caption{%
        \textbf{Overview of the evaluation scenes.}
        Plant-only Gaussian reconstructions of the five capsicum
        greenhouse-row segments used throughout our experiments.
        The collection includes short and long rows with orange-,
        red-, and yellow-fruited plants.
    }
    \label{fig:dataset_overview}
\end{figure}

\noindent
\textbf{Dataset and image acquisition.}\enspace
We evaluate Field2Sim on five segments from different planting
rows in a commercial capsicum greenhouse. Two segments contain
orange-fruited plants (\textsc{Orange-L1} and \textsc{Orange-L2}),
two contain red-fruited plants (\textsc{Red-L1} and \textsc{Red-S1}),
and one contains yellow-fruited plants (\textsc{Yellow-S1}).
We capture the images with a Samsung Galaxy S25 Ultra using
OpenCamera~\cite{harman_opencamera}, which allows us to disable the
image enhancement and post-processing typically applied by the
default camera application. Table~\ref{tab:dataset}
summarizes the capture and scene statistics, and
Fig.~\ref{fig:dataset_overview} shows the corresponding
plant-only reconstructions.

\par\noindent
\textbf{Reconstruction setup.}\enspace
Before reconstruction, we resize all images to
$1600 \times 1200$ pixels to balance reconstruction quality
against the higher memory and runtime requirements of processing
the original images. We use COLMAP~\cite{schoenberger2016sfm} to recover
camera poses and sparse scene geometry through structure from
motion, then align the reconstruction. We run all experiments
on a workstation equipped with an NVIDIA RTX A6000 GPU.

\par\noindent
\textbf{Segmentation settings.}\enspace
Unless otherwise stated, we use the same segmentation settings
across all scenes, with a scene-specific prompt for fruit
segmentation. In the prompt-selection experiments reported in
the Supplementary Material, \textit{fruit} performs best for
\textsc{Orange-L1} and \textsc{Orange-L2}, while \textit{pepper}
performs best for \textsc{Red-L1}, \textsc{Red-S1}, and
\textsc{Yellow-S1}.

\par\noindent
\textbf{Fruit instance identification.}\enspace
In the first pass, we set the proximity cutoff to
$\tau = 4.0$ Mahalanobis units and discard connected components
with fewer than 50 Gaussians. In the second pass, we set the
minimum mask support to $\gamma = 0.25$ and require evidence
from at least three views ($\nu = 3$) to flag a component for
refinement. For each flagged component, we track 12 frames and
retain tracked identities that cover at least 20\% of the
component's projected support. After splitting, we merge
fragments with fewer than 50 Gaussians into their nearest
sibling, preserving all fruit Gaussians that enter the second
pass.

\subsection{Radiance Field Reconstruction}
\label{sec:reconstruction_comparison}

We compare five 3DGS variants: the original 3DGS,
AbsGS, Mip-Splatting, Taming 3DGS, and
Hierarchical 3DGS. For the latter, we evaluate only its
chunk-level reconstruction stage and do not construct the global
hierarchy. All methods are initialized from the same SfM
reconstruction and optimized for 30,000 iterations using their
published settings, except where noted. We disable exposure
optimization for 3DGS and Hierarchical 3DGS, and use a budget multiplier of 20 for
Taming 3DGS. To ensure a consistent evaluation, we select one
eighth of the images in each scene by farthest-point sampling over
the camera centers and use the same held-out views and metric
implementation for every method. We report the unweighted mean of
the five per-scene scores.

\par\noindent
\textbf{Photometric reconstruction quality.}\enspace
Table~\ref{tab:radiance_field_reconstruction} reports PSNR,
SSIM, and LPIPS, which measure pixelwise fidelity, structural
similarity, and perceptual similarity, respectively. The metrics
do not agree on a single best method. Taming 3DGS achieves the
highest PSNR on all five scenes and the highest mean PSNR and
SSIM (21.79~dB and 0.704), whereas Hierarchical 3DGS achieves
the lowest mean LPIPS (0.262) and ranks first on LPIPS in three
scenes. Within each scene, the PSNR gap between the best and
worst methods ranges from 1.09 to 1.70~dB.\\

\begin{table}[t]
\centering
\caption{%
\textbf{Gaussian-splatting reconstruction.}
Mean novel-view synthesis performance across five scenes.
\swatch{rankFirst}\,/\,\swatch{rankSecond}\,/\,\swatch{rankThird}
highlight the best, second-, and third-best results.
$\dagger$ denotes the chunk-level optimizer of Hierarchical 3DGS.%
}
\label{tab:radiance_field_reconstruction}
\small
\setlength{\tabcolsep}{4pt}
\begin{tabularx}{\columnwidth}{@{}Xccc@{}}
\toprule
Method & PSNR\,$\uparrow$ & SSIM\,$\uparrow$ & LPIPS\,$\downarrow$ \\
\midrule
3DGS \cite{kerbl3Dgaussians}
  & \second{21.49} & \second{0.697} & \second{0.264} \\
AbsGS \cite{ye2024absgs}
  & 20.54 & 0.658 & 0.281 \\
Mip-Splatting \cite{Yu2024MipSplatting}
  & \third{21.16} & 0.683 & 0.282 \\
Taming 3DGS \cite{10.1145/3680528.3687694}
  & \first{21.79} & \first{0.704} & \third{0.265} \\
Hierarchical 3DGS$^{\dagger}$ \cite{hierarchicalgaussians24}
  & 20.72 & \third{0.688} & \first{0.262} \\
\bottomrule
\end{tabularx}
\end{table}

\par\noindent
\textbf{Downstream segmentation transfer.}\enspace
Novel-view metrics do not measure whether a rendered scene
preserves the fruit evidence required by later perception stages.
We therefore run SAM~3 with the same scene-specific prompt on
each held-out photograph and its corresponding render.
The photograph masks serve as the reference, so this evaluation
measures agreement with SAM~3 on real imagery rather than
absolute segmentation accuracy. We score only views containing
at least one reference instance, retaining renders with no
detections as complete misses.

Table~\ref{tab:reconstruction_segmentation} reports mask average
precision (AP), instance F1, Panoptic Quality (PQ), and the
missed-view rate. AP summarizes precision--recall performance,
while F1 balances instance precision and recall.
PQ~\cite{Kirillov2018PanopticS} combines detection and delineation
through $\mathrm{PQ}=\mathrm{SQ}\times\mathrm{RQ}$.
Segmentation quality (SQ) is the mean intersection-over-union
(IoU) of matched masks, while recognition quality (RQ) is an
instance F1 score that penalizes missed reference instances
and unmatched render detections. Within this evaluation, Hierarchical 3DGS achieves the strongest
measured result for every downstream metric, reaching 0.507 mask
AP, 0.655 PQ, and 0.754 F1, while reducing the missed-view rate
to 0.155. It leads on four of the five scenes, with the largest
margin on \textsc{Red-L1}, where it obtains 0.500 AP compared
with 0.355 for the next-best method. Across the five methods, mean SQ ranges narrowly from 0.862 to 0.868, indicating similar mask overlap among matched instances. The principal difference is therefore whether
fruit instances are detected, rather than the overlap of
successfully matched masks.

The downstream ranking also differs from the photometric one.
Taming 3DGS leads PSNR and SSIM but obtains 0.391 mask AP,
whereas Hierarchical 3DGS achieves the strongest measured
downstream agreement. Higher PSNR and SSIM therefore do not
necessarily correspond to better segmentation agreement within
this evaluation. Given its lowest mean LPIPS and strongest
measured downstream agreement, we use the chunk-level
Hierarchical 3DGS optimizer for the remaining Field2Sim
experiments.

 \begin{table}[t]
\centering
\caption{%
    \textbf{Downstream fruit-segmentation transfer.}
    Mean agreement between SAM~3 instance masks from held-out
    photographs and corresponding renders, using photograph
    masks as references. $\mathrm{AP}_{50:95}$ denotes mask
    average precision over IoU thresholds from 0.50 to 0.95;
    $\mathrm{F1}_{50}$ denotes instance F1 at IoU 0.5.
    ``Miss'' is the fraction of scored views with no render-side
    detection. Symbols and ranking colors follow
    Table~\ref{tab:radiance_field_reconstruction}.
}
\label{tab:reconstruction_segmentation}
\small
\setlength{\tabcolsep}{3pt}
\begin{tabularx}{\columnwidth}{@{}Xcccc@{}}
\toprule
Method & AP$_{50:95}\uparrow$ & PQ$\uparrow$ & F1$_{50}\uparrow$ & Miss$\downarrow$ \\
\midrule
3DGS \cite{kerbl3Dgaussians}
  & \third{0.400} & \third{0.582} & \third{0.673} & \third{0.235} \\
AbsGS \cite{ye2024absgs}
  & \second{0.426} & \second{0.606} & \second{0.702} & \second{0.212} \\
Mip-Splatting \cite{Yu2024MipSplatting}
  & 0.392 & 0.566 & 0.652 & 0.275 \\
Taming 3DGS \cite{10.1145/3680528.3687694}
  & 0.391 & 0.569 & 0.655 & 0.270 \\
Hierarchical 3DGS$^{\dagger}$ \cite{hierarchicalgaussians24}
  & \first{0.507} & \first{0.655} & \first{0.754} & \first{0.155} \\
\bottomrule
\end{tabularx}
\end{table}

\subsection{Fruit Instance Identification}
\label{sec:instance_evaluation}

We evaluate fruit instance identification on 1,722 fruit
instances comprising 363,844 Gaussian primitives across the
five scenes. Ground-truth instance labels were manually
created for evaluation. All methods are evaluated on the same
reference fruit Gaussians, allowing predicted and reference
instances to be compared through exact set intersections.

We compare our complete two-pass method against
DBSCAN~\cite{3001460.3001507} and HDBSCAN \cite{campello_density-based_2013}, as well as an
ablation containing only the first geometric pass.
All methods use a minimum instance size of 50 Gaussians,
with baseline distance parameters normalized by the median
distance to the tenth-nearest fruit Gaussian in each scene.
We report Panoptic Quality (PQ) at an IoU threshold of 0.5
and its segmentation and recognition components (SQ and RQ).
We also report Adjusted Rand Index (ARI), which measures
agreement between predicted and reference instance assignments
over pairs of Gaussians, adjusted for chance.
Absolute relative count error measures the absolute difference
between predicted and reference instance counts as a percentage
of the reference count.

\par\noindent
\textbf{Quantitative results.}\enspace
Table~\ref{tab:instance_identification} reports the unweighted
mean over the five scenes. The complete method reaches
0.954 PQ and 0.963 ARI, compared with 0.808 PQ and 0.789 ARI
for the geometric pass alone. The second pass therefore
improves PQ by 0.146 and reduces mean absolute relative
count error from 20.0\% to 1.5\%.
Across all five scenes, the number of reference instances
matched to a prediction increases from 1,354 to 1,667 out
of 1,722, while unmatched predictions increase from 42 to 51.
The complete method achieves PQ scores between 0.943 and
0.976 across scenes.

The geometric pass achieves PQ comparable to the clustering
baselines, with 0.808 compared with 0.785 for HDBSCAN and
0.774 for DBSCAN. Both density-based methods under-count
in every scene, with mean absolute relative count errors
of 20.5\% and 12.4\%, respectively. Together with the
geometric-pass ablation, these results indicate that the
image-driven refinement provides the main improvement in
instance identification.

\begin{table}[t]
    \centering
    \caption{%
        \textbf{Fruit instance identification.}
        Unweighted means over five scenes.
        Error denotes absolute relative instance-count error.
        Best results are shown in bold.
    }
    \label{tab:instance_identification}
    \footnotesize
    \setlength{\tabcolsep}{3pt}
    \renewcommand{\arraystretch}{1.08}
    \begin{tabular*}{\columnwidth}
        {@{\extracolsep{\fill}}lccccc@{}}
        \toprule
        Method &
        PQ~$\uparrow$ &
        SQ~$\uparrow$ &
        RQ~$\uparrow$ &
        ARI~$\uparrow$ &
        Error (\%)~$\downarrow$ \\
        \midrule
        DBSCAN &
            0.774 & 0.904 & 0.857 & 0.591 & 12.4 \\
        HDBSCAN &
            0.785 & 0.929 & 0.845 & 0.765 & 20.5 \\
        \midrule
        Ours (Geometric) &
            0.808 & 0.944 & 0.856 & 0.789 & 20.0 \\
        Ours (Two-pass) &
            \textbf{0.954} &
            \textbf{0.985} &
            \textbf{0.969} &
            \textbf{0.963} &
            \textbf{1.5} \\
        \bottomrule
    \end{tabular*}
\end{table}

\subsection{Practical Utility}
\label{sec:practical_utility}

Field2Sim produces reusable visual simulation assets from
handheld smartphone images without specialized capture equipment.
Across the five captures, a single smartphone was sufficient to
reconstruct 23.4\,m of greenhouse rows containing approximately
55 plants and 1,722 fruits. Each scene contains a conventional
3DGS model, plant and fruit subsets, and individual fruit labels.
These assets can be rendered using standard Gaussian rasterizers
without additional neural parameters.

\begin{figure}[t]
    \centering
    \includegraphics[
        width=\columnwidth
    ]{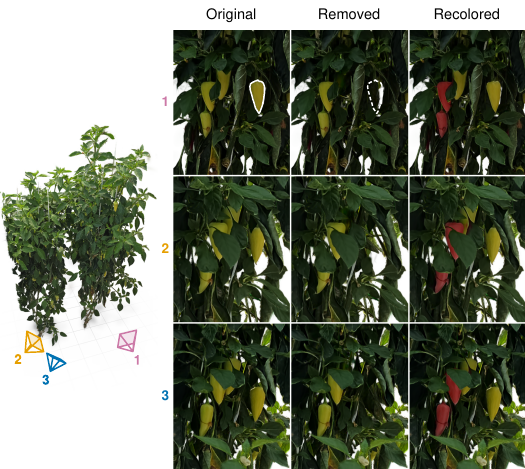}
\caption{%
    \textbf{Instance-level editing for visual simulation.}
    Left shows the reconstructed plant asset and three camera
    viewpoints. Right shows the original, fruit-removal, and
    recolored configurations from each viewpoint.
    Solid and dashed outlines in the first row mark an example
    fruit and its location after removal.
    Removal varies fruit count and spatial distribution,
    while recoloring varies visual cues associated with ripeness.
}
    \label{fig:practical_utility}
\end{figure}

Separating individual fruits allows scene configurations to be
varied without repeating capture or reconstruction.
Figure~\ref{fig:practical_utility} demonstrates fruit removal
and recoloring from three viewpoints. Removing different subsets
of fruits produces many configurations with different counts
and spatial distributions while preserving the underlying plant
structure. Recoloring changes the visual cues associated with
ripeness, allowing scenarios with different distributions of
visually ripe target fruits. Fruits can also be duplicated or
repositioned to modify their arrangement. Because these edits
are applied to the shared 3D representation, each configuration
is consistent across rendered viewpoints by construction.

The resulting configurations support fruit-counting and
segmentation evaluation, as well as information-gathering and
view-planning tasks that seek to discover fruits or inspect
targets with particular appearance attributes. RGB images,
depth maps, and corresponding semantic or instance masks can
be rendered from camera poses within the reconstructed view
support. This allows camera placement and perception-driven
view selection to be evaluated across controlled scene
variants without repeatedly accessing the greenhouse.
Since annotation is performed post hoc, updated segmentation
models or labels for additional plant organs can also be
incorporated without reconstructing the scene again.

The assets are intended for visual simulation and perception. They do not provide watertight collision geometry, biomechanical properties, or plant dynamics, and therefore do not yet support contact-rich tasks such as harvesting or leaf manipulation, which we leave to future work.

\section{Conclusions}
\label{sec:conclusions}

Field2Sim demonstrates that handheld smartphone images can yield
photorealistic plant assets with individual fruit labels for visual
simulation. Our evaluation highlights two considerations for building
such assets: reconstruction quality should be assessed through
downstream perception as well as image similarity, and multi-view
instance evidence helps separate neighbouring fruits that geometric
clustering alone leaves merged. Together, these findings support
combining established reconstruction and segmentation methods with
targeted instance refinement to produce reusable, editable crop assets.

The resulting assets provide a basis for controlled studies of fruit
perception and view planning, with fruit configurations that can be
varied without additional capture. Our evaluation is limited to static
capsicum rows from one greenhouse; extending it to other crops and
testing the assets in closed-loop robotic tasks are the next steps
toward establishing their broader utility.
\vfill

\section*{Acknowledgements}
This paper was made possible by NPRP grant 14S-0417-210227 from the Qatar Research, Development and Innovation (QRDI) Council / Qatar National Research Fund (QNRF). The statements made herein are solely the responsibility of the authors.

{
    \small
    \bibliographystyle{ieeenat_fullname}
    \bibliography{main}
}

\clearpage
\maketitlesupplementary

\section{Effect of Prompt Choice}
\label{sec:text_prompts}

As discussed in Sec.~\ref{sec:semantic}, SAM~3
\cite{carion2025sam3segmentconcepts} accepts a text prompt rather than a fixed
semantic category, making the choice of noun a potential source of variation.
We evaluate four candidate prompts on the same 100 sampled frames from each
scene, for 500 frames in total. All prompts are evaluated using identical
inference settings. A frame is counted as detected when SAM~3 returns at least
one predicted instance. Mask coverage is computed as the fraction of image
pixels contained in the union of the predicted masks, averaged over the sampled
frames. These measures characterize the model's response to a prompt; they do
not constitute an accuracy evaluation against ground truth.

\begin{figure}[t]
    \centering

    \begin{subfigure}[t]{\linewidth}
        \centering
        \includegraphics[
            width=\linewidth
        ]{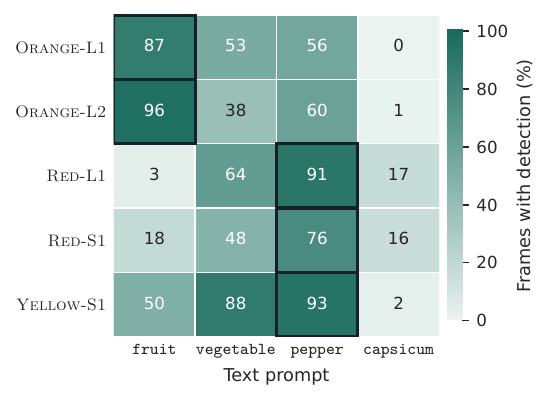}
        \caption{Detection frequency.}
        \label{fig:prompt-effect-detection}
    \end{subfigure}

    \vspace{0.5em}

    \begin{subfigure}[t]{\linewidth}
        \centering
        \includegraphics[
            width=\linewidth
        ]{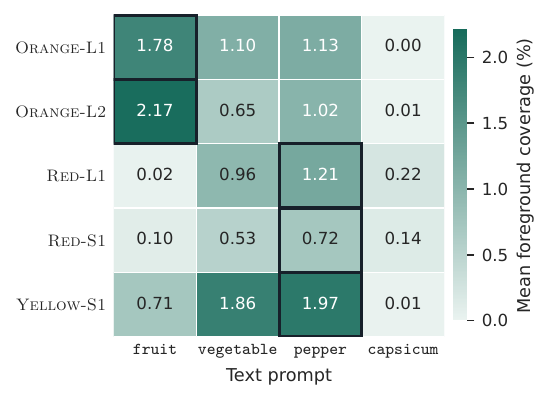}
        \caption{Predicted mask coverage.}
        \label{fig:prompt-effect-coverage}
    \end{subfigure}

    \caption{
        Effect of prompt choice across the five evaluation scenes.
        Each cell is computed from the same 100 sampled frames.
        The outlined cell denotes the largest response within a scene
        and metric. The prompt \texttt{fruit} produces the strongest
        response in the orange scenes, whereas \texttt{pepper} is
        stronger in the red and yellow scenes.
    }
    \label{fig:prompt-effect}
\end{figure}

\begin{figure*}[t]
    \centering
    \setlength{\tabcolsep}{2pt}
    \renewcommand{\arraystretch}{0.7}

    \begin{tabular}{
        @{}
        >{\raggedright\arraybackslash}m{0.115\textwidth}
        cccc
        @{}
    }
        &
        \texttt{fruit} &
        \texttt{vegetable} &
        \texttt{pepper} &
        \texttt{capsicum}
        \\[4pt]

        \textsc{Orange-L2} &
        \includegraphics[
            width=0.205\textwidth
        ]{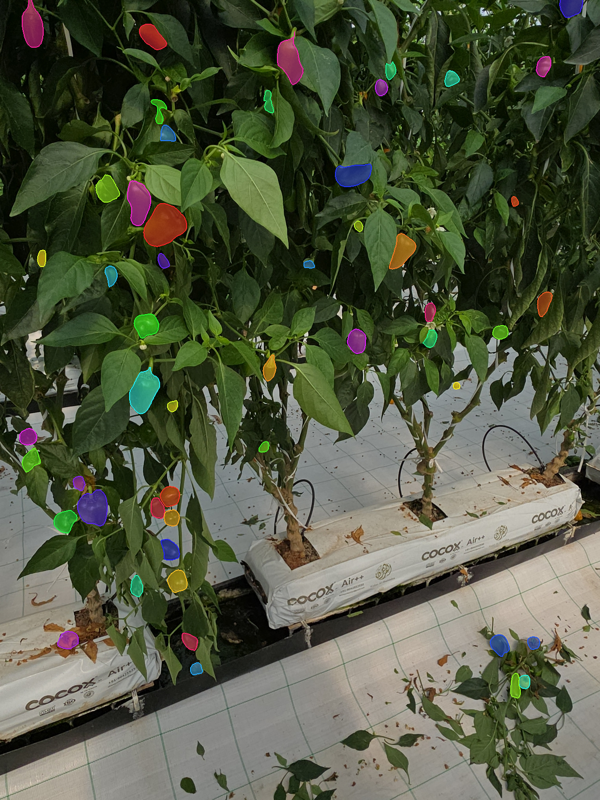} &
        \includegraphics[
            width=0.205\textwidth
        ]{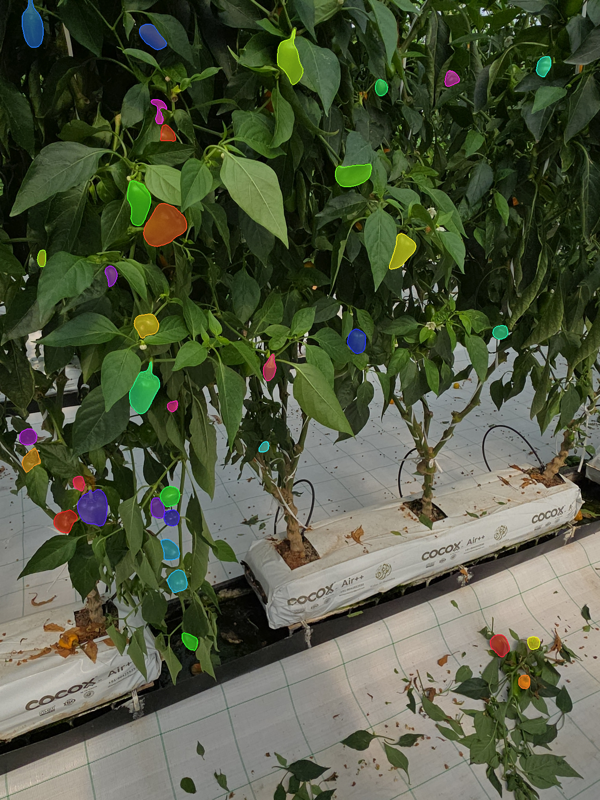} &
        \includegraphics[
            width=0.205\textwidth
        ]{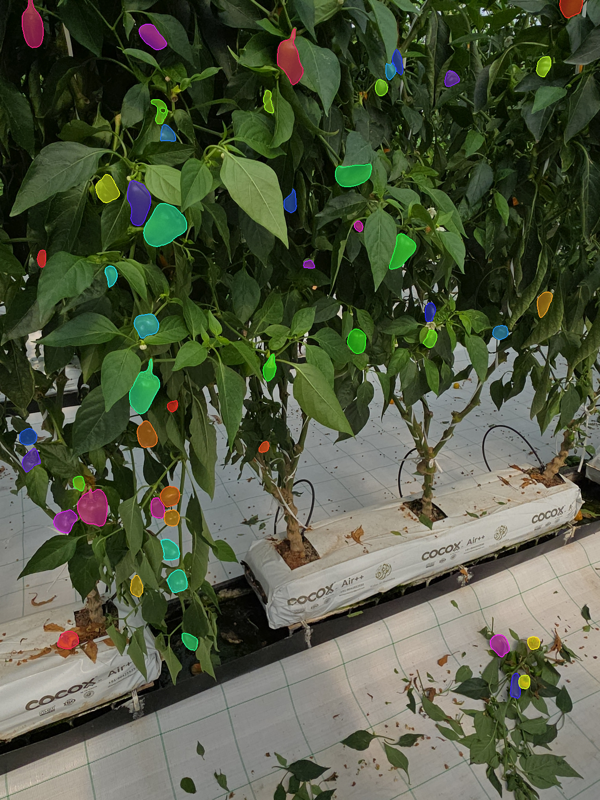} &
        \includegraphics[
            width=0.205\textwidth
        ]{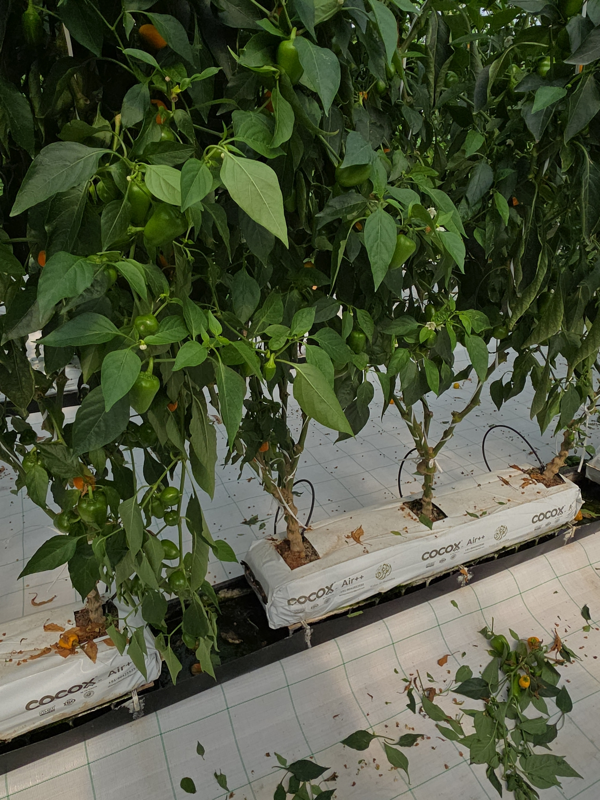}
        \\[3pt]

        \textsc{Red-S1} &
        \includegraphics[
            width=0.205\textwidth
        ]{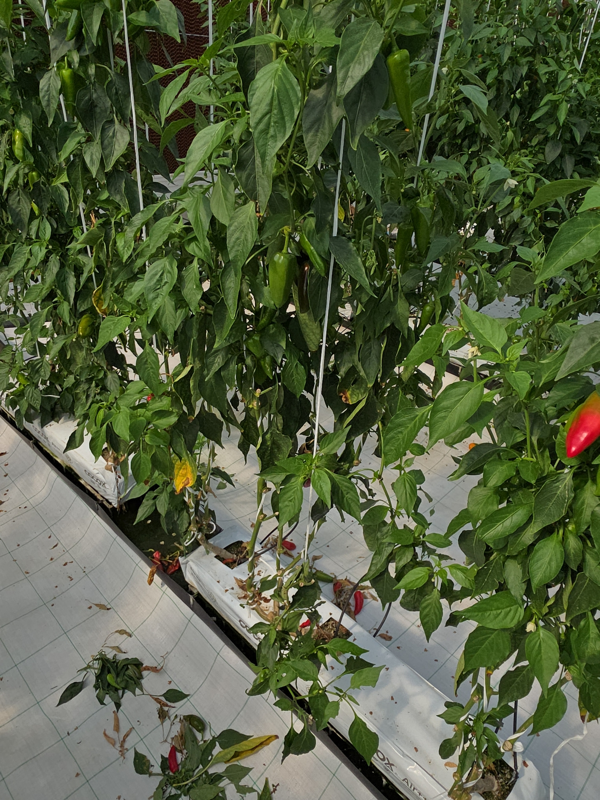} &
        \includegraphics[
            width=0.205\textwidth
        ]{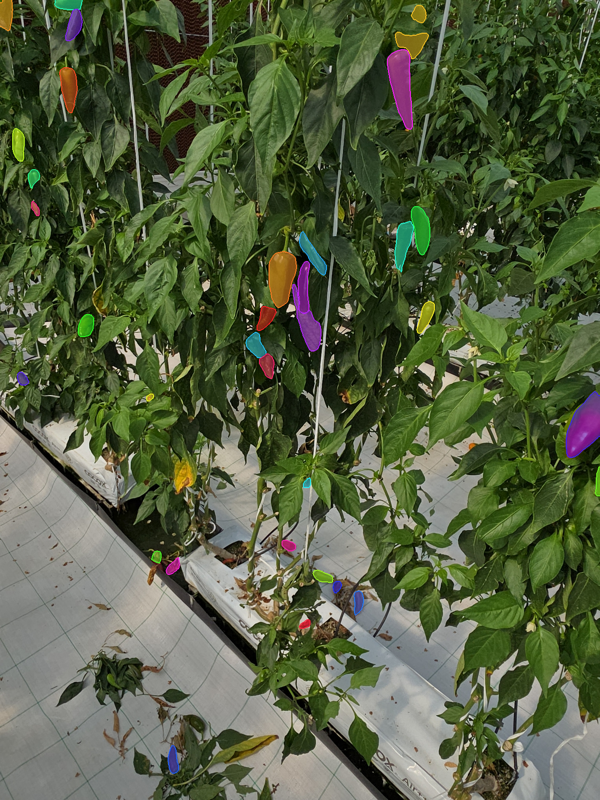} &
        \includegraphics[
            width=0.205\textwidth
        ]{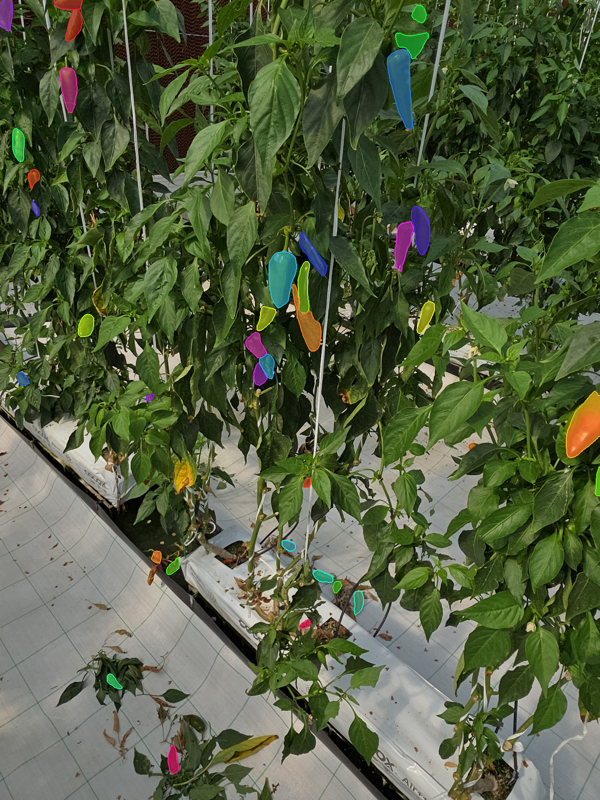} &
        \includegraphics[
            width=0.205\textwidth
        ]{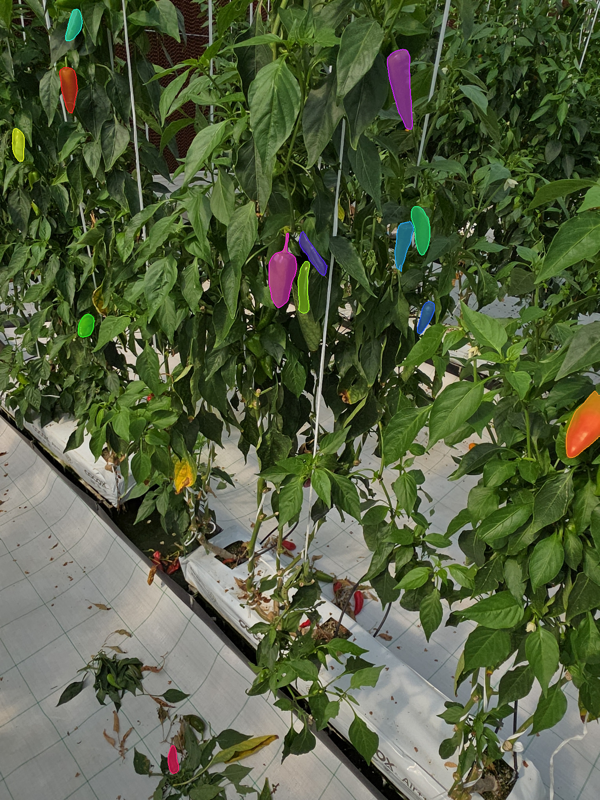}
        \\[3pt]

        \textsc{Yellow-S1} &
        \includegraphics[
            width=0.205\textwidth
        ]{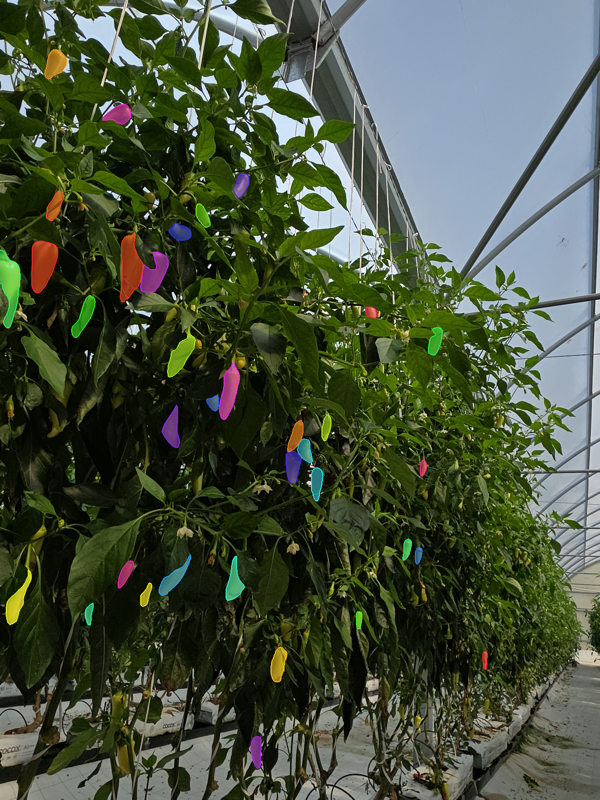} &
        \includegraphics[
            width=0.205\textwidth
        ]{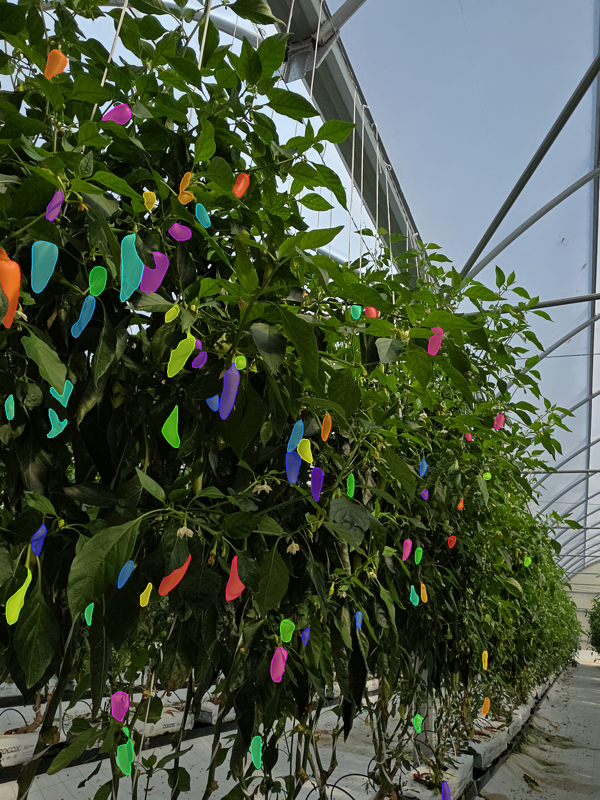} &
        \includegraphics[
            width=0.205\textwidth
        ]{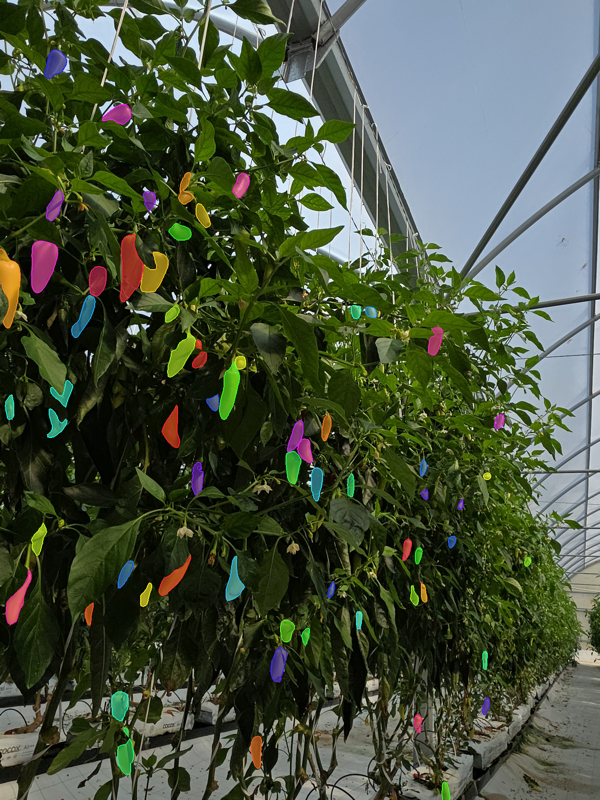} &
        \includegraphics[
            width=0.205\textwidth
        ]{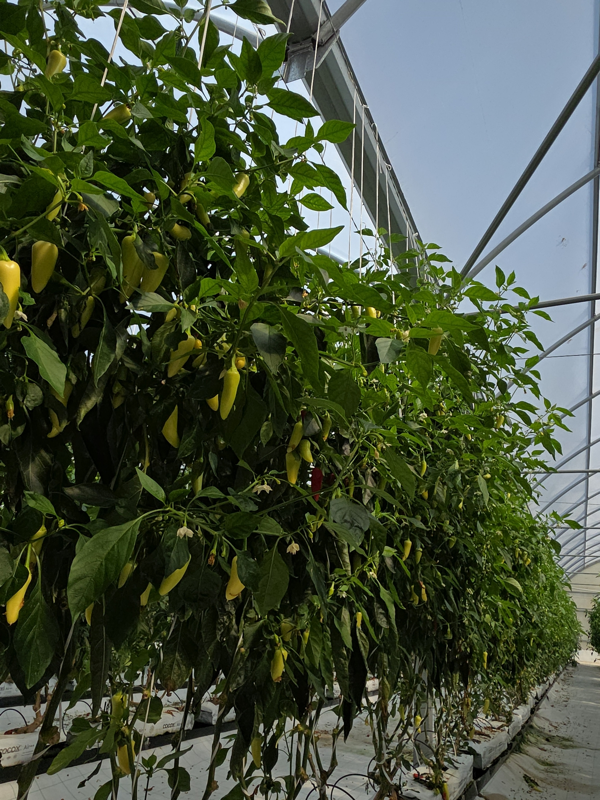}
    \end{tabular}

    \caption{
        Qualitative segmentation results for different text prompts.
        Each column corresponds to one prompt, while each row shows
        predictions for the same representative view from
        \textsc{Orange-L2}, \textsc{Red-S1}, and \textsc{Yellow-S1},
        respectively. Colours distinguish predicted instances within
        each image and do not indicate correspondence across prompts.
    }
    \label{fig:prompt_comparison}
\end{figure*}

Figure~\ref{fig:prompt-effect} shows that the effect is strongly scene
dependent. The prompt \texttt{fruit} yields detections in 87\% and 96\% of
frames in \textsc{Orange-L1} and \textsc{Orange-L2}, respectively, but only
3\% and 18\% in the two red scenes. Conversely, \texttt{pepper} reaches 91\%
and 76\% on the red scenes and 93\% on \textsc{Yellow-S1}, while also producing
the largest mask coverage in these three scenes. The term \texttt{capsicum}
rarely activates the model, with an aggregate detection rate of only 7.2\%.
Across all 500 frames, \texttt{pepper} provides the highest aggregate detection
rate and mean coverage, at 75.2\% and 1.21\%, respectively. Examples can be seen
in \cref{fig:prompt_comparison}.

For each scene, we select the prompt with the highest detection rate. The same
prompt also maximizes mean mask coverage in all five scenes. This selects
\texttt{fruit} for \textsc{Orange-L1} and \textsc{Orange-L2}, and
\texttt{pepper} for \textsc{Red-L1}, \textsc{Red-S1}, and
\textsc{Yellow-S1}. The reversal between the orange and remaining scenes shows
why a single fixed prompt is not reliable across captures.

\section{Qualitative Visualizations}
\label{sec:qualitative-clustering}

Figure~\ref{fig:clusters-red1} presents qualitative examples of the instance
clustering results. These crops illustrate cases where purely geometric
clustering merges nearby fruits, while the second pass recovers the individual
instances.

\newcommand{\clusterpanel}[1]{%
    \includegraphics[width=\linewidth]{#1}%
}
\graphicspath{{fig/clusters/}}

\begin{figure*}[t]
    \centering
    \setlength{\tabcolsep}{3pt}
    \renewcommand{\arraystretch}{1.0}

    \begin{tabularx}{\textwidth}{
        @{}
        *{5}{>{\centering\arraybackslash}X}
        @{}
    }
        \shortstack[c]{
            \small\textbf{Ground Truth}\\[-1pt]
            \scriptsize Reference
        }
        &
        \shortstack[c]{
            \small\textbf{Ours}\\[-1pt]
            \scriptsize Two-pass
        }
        &
        \shortstack[c]{
            \small\textbf{Ours}\\[-1pt]
            \scriptsize Geometry only
        }
        &
        \shortstack[c]{
            \small\textbf{DBSCAN}\\[-1pt]
            \scriptsize Baseline
        }
        &
        \shortstack[c]{
            \small\textbf{HDBSCAN}\\[-1pt]
            \scriptsize Baseline
        }
        \\[5pt]

        \clusterpanel{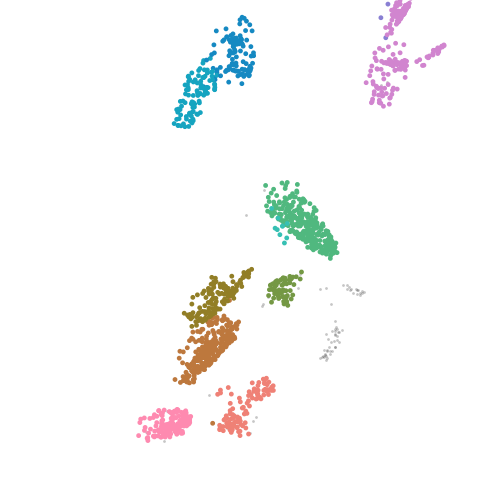} &
        \clusterpanel{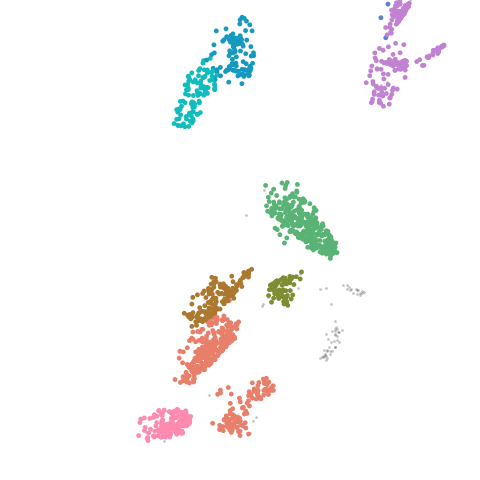} &
        \clusterpanel{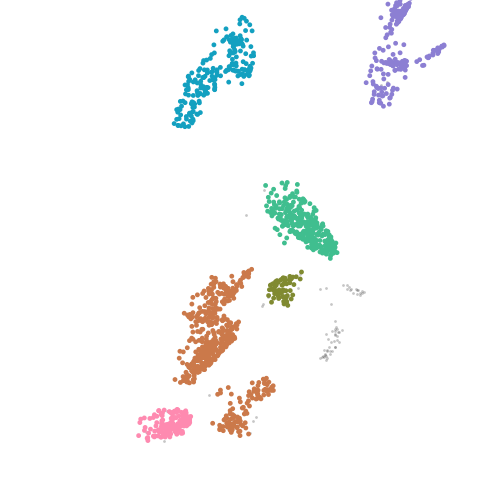} &
        \clusterpanel{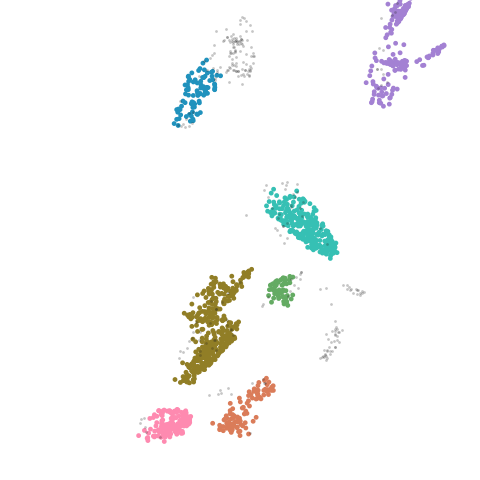} &
        \clusterpanel{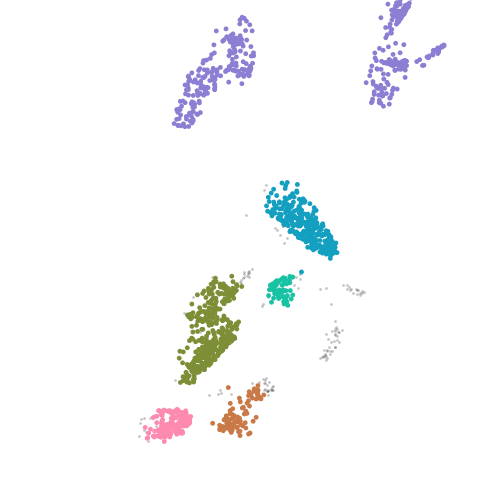}
        \\[4pt]

        \clusterpanel{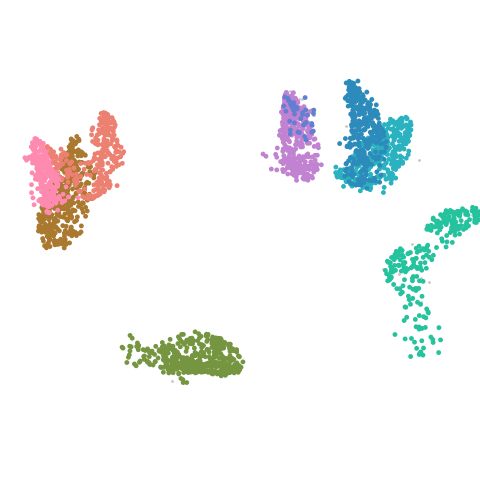} &
        \clusterpanel{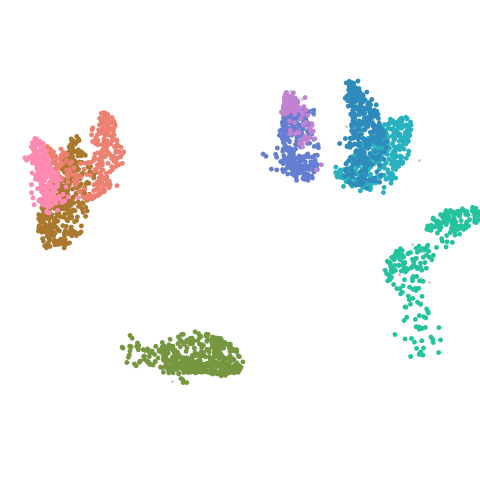} &
        \clusterpanel{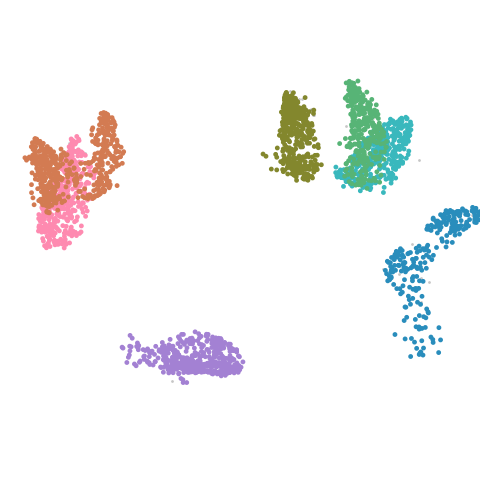} &
        \clusterpanel{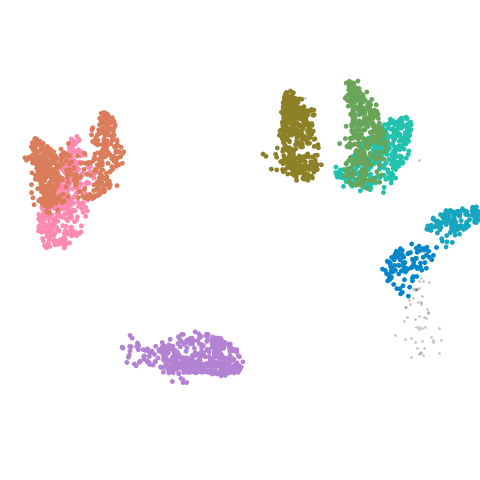} &
        \clusterpanel{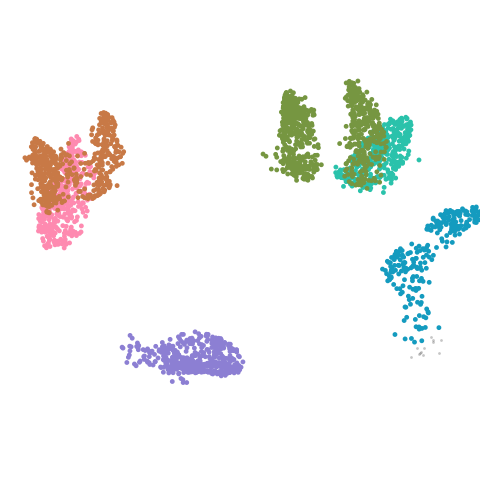}
        \\[4pt]

        \clusterpanel{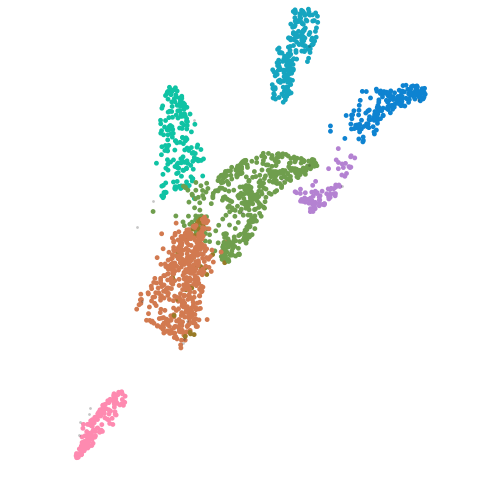} &
        \clusterpanel{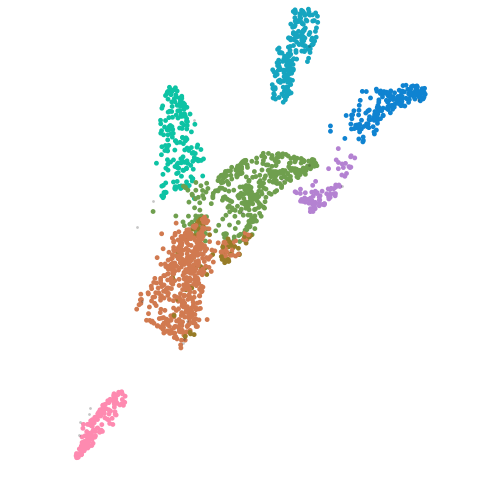} &
        \clusterpanel{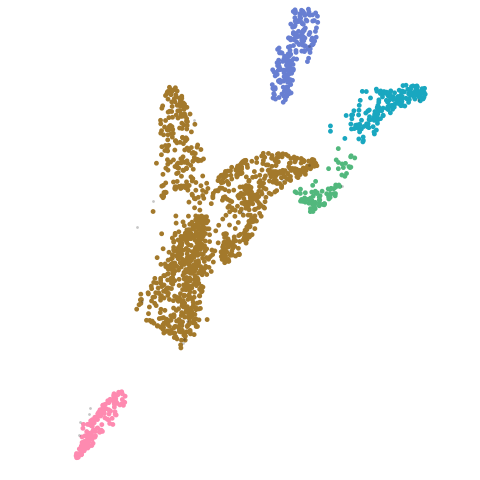} &
        \clusterpanel{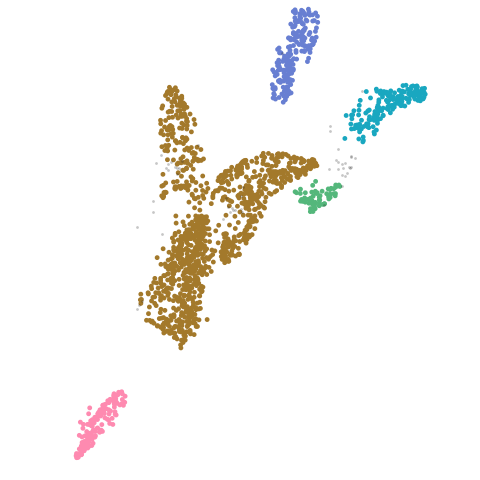} &
        \clusterpanel{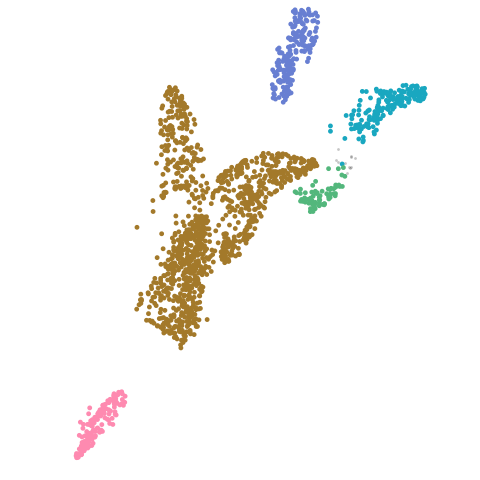}
    \end{tabularx}

    \caption{
        Qualitative instance-clustering results on
        \textsc{Red-L1}. The three non-overlapping crops are selected
        from regions where the methods disagree most, measured by their mean
        \(1-\mathrm{IoU}\) against the ground-truth instances. Every method
        partitions the same set of object Gaussians. Colours are assigned
        independently within each panel and therefore do not correspond across
        columns; small grey points denote unassigned Gaussians. The two-pass
        method separates nearby instances that remain merged under purely
        geometric clustering. Scene-level panoptic quality is \(0.950\) for
        ours, \(0.824\) for geometry only, \(0.782\) for DBSCAN, and \(0.786\)
        for HDBSCAN.
    }
    \label{fig:clusters-red1}
\end{figure*}

\end{document}